# AI Models Close to your Chest: Robust Federated Learning Strategies for Multi-site CT

Edward H. Lee[1,*], Brendan Kelly[2], Emre Altinmakas[3], Hakan Dogan[4], Maryam Mohammadzadeh[5], Errol Colak[6], Steve Fu[7], Olivia Choudhury[7], Ujjwal Ratan[7], Felipe Kitamura[8], Hernan Chaves[9], Jimmy Zheng[1], Mourad Said[10], Eduardo Reis[11], Jaekwang Lim[12], Patrícia Yokoo[11], Courtney Mitchell[13], Golnaz Houshmand[14], Marzyeh Ghassemi[15], Ronan Killeen[2], Wendy Qiu[13], Joel Hayden[13], Farnaz Rafiee[14], Chad Klochko[16], Nicholas Bevins[16], Faeze Sazgara[17], S. Simon Wong[18], Michael Moseley[1], Safwan Halabi[19], Kristen W. Yeom[1,*].

[1]School of Medicine, Stanford University, USA
[2]Insight Centre for Data Analytics, University College Dublin, Ireland
[3]Icahn School of Medicine at Mount Sinai, USA
[4]Koç University School of Medicine, Turkey
[5]Division of Neuroradiology, Amiralam Hospital, Tehran University of Medical Sciences, Iran
[6]Unity Health Toronto, St. Michael's Hospital and University of Toronto, Canada
[7]Amazon Web Services
[8]Universidade Federal de São Paulo, Brazil
[9]Fleni, Argentina
[10]Radiology Department Centre International Carthage Médicale, Monastir, Tunisia
[11]Hospital Israelita Albert Einstein, São Paulo, SP, Brazil
[12]Department of Radiology, Kyungpook National University Hospital, Korea
[13]Department of Radiology, St. Joseph's Hospital and Medical Center, USA
[14]Rajaie Cardiovascular Medical and Research Center, Iran University of Medical Sciences
[15]Dept. of Electrical Engineering and Computer Science, at MIT, USA
[16]Department of Radiology, Henry Ford Health System, Detroit MI
[17]Department of Radiology, Guilan University of Medical Sciences, Iran
[18]Department of Electrical Engineering, Stanford University, USA
[19]Ann & Robert H. Lurie Children's Hospital of Chicago

*Co-corresponding authors

**ABSTRACT**

*While it is well known that population differences from genetics, sex, race, and environmental factors contribute to disease, AI studies in medicine have largely focused on locoregional patient cohorts with less diverse data sources. Such limitation stems from barriers to large-scale data share and ethical concerns over data privacy. Federated learning (FL) is one potential pathway for AI development that enables learning across hospitals without data share. In this study, we show the results of various FL strategies on one of the largest and most diverse COVID-19 chest CT datasets: 21 participating hospitals across five continents that comprise >10,000 patients with >1 million images. We also propose an FL strategy that leverages synthetically generated data to overcome class and size imbalances. We also describe the sources of data heterogeneity in the context of FL, and show how even among the correctly labeled populations, disparities can arise due to these biases.*

**INTRODUCTION**

AI systems in natural images and language processing have thrived on the availability of large data that dwarf those in medicine. Medical data, however, are often less accessible, siloed across hospitals, and require ethical board and legal approval in observation of government regulations such as Health Insurance Portability and Accountability Act (HIPAA) and the European General Data Protection Regulation (GDPR) that are intended to serve as guardrails to protect patient privacy and safety. Unfortunately, data scarcity in medicine, both in *size* and *diversity*, places upper bounds on the size, complexity, and robustness of clinical AI models. Centralized data sharing (CDS) among a few siloed institutions may boost generalization performance [1-13], but regulatory barriers to CDS can delay research and real-time advances of AI in medicine.

In addition to data *sparsity,* general lack of *data diversity* in AI research in medicine, remains a fundamental problem. Rare diseases are difficult to aggregate, dispersed across hospitals; and imbalances in disease distribution or phenotype differences due to demographics, genetics, race, disease variants, and environmental or other factors, including data acquisition methods (e.g., laboratory toolkits, hardware, protocols), can introduce unknown biases across geographic centers that can impede meaningful AI developments.

Collaborative learning could offer an alternative path for large-scale data exploration and site-to-site interactive learning that advance the goals of high model performance and site-specific generalization, while respecting the privacy of siloed patient data. For example, Federated Learning (FL) by McMahan et al. [14] is one collaborative method that uses AI training without data share, whereby each collaborative site within an FL network agrees on one training model and optimization problem, then shares information of its weights (or gradients) during the training process. In contrast, conventional CDS method uses pooled data shared from individual sites. The most common FL approach is Federated Averaging (FedAvg). In FedAvg, AI models work locally on individual site, including site-specific local gradient updates and weight transfer back to a central server. The weights from all sites are then averaged (reduced) and subsequently broadcasted back to individual sites for the next round of training. The number of local gradient updates per communication (or aggregation) round is an important FL-specific hyperparameter.

Other collaborative techniques include fine-tuning, commonly used when the weights trained from another site are accessible, e.g., from a public dataset. Typically, for best generalization performance, the public dataset larger than the sites used to fine-tune is desirable as the large dataset helps to capture shared feature representations. A model can also be transferred from site 1, fine-tuned on site 2, and this process repeated for a chain of multiple sites. For example, in Institutional Incremental Learning (IIL) and Cyclical Institutional Incremental Learning (CIIL), one model traverses one site to another for incremental weight updates [17,18]. Unlike FedAvg, IIL or CIIL does not require a central server for synchronization and communication from individual sites.

In general, however, communicating weights across a large number of sites with high number of local gradient updates per round can tax training and incur additional overhead. If one site has few data samples while other sites have 100 times the size, issues of global convergence can also arise for all collaborative methods.  For example, in IIL and CIIL, one slow site in the network chain can impair training on the entire network, whereas asynchronous [19] methods in FedAvg can mitigate such bottlenecks. Conceptually, FedAvg is also likely more resilient against sites that contain adversaries which, knowingly or unknowingly, could inject corrupt or skewed data or stall the network.

Recent studies have shown feasibility of FL in medicine using COVID chest x-ray [15] and 3D brain MRI for tumor segmentation [16]. Given diverse and evolving nature of medical data that stem from interventions, mutations, and patient-specific or environmental risk factors, these preliminary studies prompt investigations regarding FL robustness on data heterogeneity and population subgroups. Works by [33-36] offer state-of-the-art FL solutions in healthcare and address the open challenge with FL on skewed data and class imbalance.

Recent surges and high transmissibility of COVID-19 (COVID) have highlighted a need for fast and accurate model developments with capability for real-time updates and efficiency in medicine, including preparedness for future unknown pandemics. Here, we specifically target large and heterogeneous COVID and control Chest CT datasets from multi-national geographic centers and across multiple time points, enriched with infectious variants and diverse disease severity, age groups, and ethnic populations (**Table 1**).  We examine promising collaborative learning methods (**Figure 1**) on 3D chest CT lung evaluation and assess its robustness in the context of heterogeneous data distribution. We also examine site-specific FL performances as well as within each sex and age subgroups and potential techniques to mitigate class imbalances.

**CONTRIBUTIONS**

First, we show proof-of-concept FL algorithms on an international 3D Chest CT dataset for COVID prediction. To the best of our knowledge, the study represents the largest and most diverse COVID chest CT datasets: 16 sites (21 participating hospitals) across five continents that comprise >10,000 patients with >1 million images. Second, we highlight the challenges of training with real world data that are diverse and imbalanced, or non-independent and identically distributed (non-IID). For example, with siloed training, we show how generating synthetic data conditioned on the site of interest can boost performance on data-sparse or class-imbalanced sites. We show that due to heterogeneous data compositions across sites, CIIL and IIL methods underperform compared to FL should not go unnoticed.

**RESULTS**

**Baseline siloed training and techniques for model augmentation**
*Baseline siloed model development.*  We present 20 baseline models (2.1-16.1 datasets) trained on local datasets from individual sites (Sites 2-16), using the same 3D architecture (see **Methods**). All 20 models are evaluated on the hold-out test sets from each respective site (Sites 2-16) plus a hold-out test set from Site 1 public dataset (1.1), for a total of 20 x 21 evaluations. We find that the best siloed model achieves a 65% per site weighted average accuracy. There were some sites that had missing examples from one of the classes. In the per-site analysis, we choose to report the accuracies instead of F1-scores as F1-scores would be ill-defined for the missing classes. In our aggregated accuracies (Figure 3 (c)), we report the weighted-average F1-score for the entire validation set. The matrix showing the 20 x 21 validations are shown in **Figure S2(a)**. Here, we see a wide range in model performance. First, most models trained on individual sites show lackluster performance when tested on Site 1 [3], possibly due to a unique image source from Site 1 (non DICOM images with only segmented lung, i.e., no chest wall or heart). Second, a few of the sites have small samples, i.e., <100 patients; and the siloed models built from these sites show low performance (weighted average accuracy). Third, larger sites do not always equate with the highest performance.

*Pretraining*. If data intermixing across sites is not feasible, pretraining models is a logical alternative for siloed training. Pretraining on a large public data could capture shared feature representations in the CT chest domain, potentially improve bias-variance tradeoff, and help build generalizable models with higher complexity and dimensionality. Hence, we add pretraining using a large public dataset (1.1) to our siloed experiment. Per site weighted average and average accuracies rise significantly higher (**Figure S1(b)**) compared to earlier baseline model without pretraining (**Figure S1(a)**). However, even with pretraining, some sites continue to display poor performance.

*Synthetic data generation*. Next, we present a GAN-based approach for image synthesis, that could potentially off-set class imbalance, applicable to both siloed and FL training. This method is particularly relevant to FL, since learning with skewed class proportions *and* non-IID data compositions both present fundamental problems in FL. We show that a GAN-generator can augment the training data of the underrepresented classes by synthesizing representative samples from a conditional GAN trained only on the public dataset 1.1 and conditioned on the site of interest during the inference-time. Our GAN-generator takes as input any real CT image from the site of interest plus corresponding style label, then generates synthetic CT image labels (including both lung and external tissues) that are representative of sites. Examples are shown in **Figure 2**. We show that this method can effectively equalize the distribution of classes across all sites during training, while nearly doubling the overall training dataset. This can be considered an extension of pretraining using 1.1 data (**Figure S1**) and also does not affect the hold-out test sets used for evaluation.

**Federated Learning Strategies**
**Figure 3** summarizes the CDS and FL results from the test sets from all sites (Sites 1-16, or 1.1-16.1, totaling 21 datasets). All CDS and FL experiments use the same 3D model architecture from the earlier siloed training (see **Methods**). With 1 local epoch per reduction round, we observe 0.78 average and 0.80 weighted average accuracies for the 21 models across all sites. Training with CDS performs the highest as the global model has access to all data. FL training *only with synthetic data* from each site (2.1-16.1 datasets) yields surprising results, i.e., higher performance than any siloed strategy. FL training using *both synthetic and real data* for each site (2.1-16.1 datasets) outperforms synthetic data-only FL and performs on-par using all site data (1.1-16.1 datasets). In fact, we find that in many sites that are data-sparse or class-imbalanced, synthetic data can boost accuracies. In summary, we find that synthetic data *in conjunction with real data* can slightly help with out-of-distribution datasets. However, it is unclear whether this method can be generalized for all FL datasets. Nonetheless, synthetic data generation could facilitate data share, help mitigate imbalances across sites, and potentially offer quality control among collaborators as they can view representative scans between sites (**Figure 2**).

**Alternative Methods to Federated Learning**
We evaluate alternatives to FL, such as IIL and CIIL. **Figure 3 (b)** shows site-specific performances of IIL and CIIL across different training paths. Due to the heterogeneity of our dataset, the *choice of training path* dramatically impacts final precision. For example, IIL that traverses from sites with the smallest to the largest cohort shows more robust accuracies than IIL going from the largest to the smallest. Some sites can also be considered out-of-distribution (e.g., lack of PNA samples) or have highly skewed distributions (e.g., few data samples, non DICOM data sources, segmented lung without chest wall), and thus IIL and CIIL may be less suitable than FL for learning. We find that IIL and CIIL methods tend to perform similarly. When training over the smallest cohort-site (Site 14), validation accuracies deteriorate regardless of the path trajectory.

If, however, we artificially partition the training datasets to be independent and identically distributed (IID) and balanced in the number of training sample, we find that CIIL performance improves (**Figure 3b**, top row) with accuracies that are fairly robust even with 3 rounds, and significantly higher than the CIIL performed on non-IID datasets. The IID data experiments suggest vulnerability of IIL and CIIL methods with heterogeneous datasets although further research is warranted.

**Robustness of Federated Learning on non-IID data**
Data distribution across individual sites can impact the training convergence and final accuracy of the federated model [14, 23]. Thus, we explore the influence of IID data, or lack thereof, on FL training. Non-IID scenarios might include discrepant disease severity, variable gray-scale mapping or CT Hounsfield Units, missing or imbalanced classes, and whole chest images versus segmented lung, typically found in a real-world setting. **Figure 4 (a)** illustrates the impact of such highly skewed non-IID data in the Federated network across all sites. Each site performs 1 epoch of local training and sends the trained weights to a parameter server to be averaged via a reduction round. In the non-IID case, each site represents an independent hospital. In the simulated IID scenario, we aggregate all training data from all sites, randomly shuffle, and partition the data with equal sample sizes without replacement across the sites. We see that convergence for IID is overall smoother and higher (a). In **Figure 4(d)**, IID achieves higher final accuracies than non-IID. A FL-specific hyperparameter is the fraction of sites that perform computation on each reduction round, denoted as "C" in **Figure 4**. We observe that the choice of C matters more in the non-IID, high diversity scenario. In IID, we find little effect in the choice of C.

We examine different model architectures and find that the choice of model architecture (ResNet-3D, Inception-3D, etc.) does not significantly affect performance. However, the data pipeline does significantly affect FL. We compare 2D MobileNet (ImageNet-pretrained) and 3D Inception (Kinetics-pretrained) CNNs [24-25]. The 3D models tend to achieve much higher performance than 2D in both the IID and non-IID scenarios **(Fig. 4(b))**. Pipelines restricted to only 2D slices fails to fully capture subtle COVID or PNA feature variations.

We find that altering the number of participating sites can also alter convergence. In **Figure 4 (d)**, we artificially split each site-specific datasets (N=21) into two or three subsets, while keeping the total sample size the same. In the IID scenario, the performance remains equivalent across the 21 (baseline) versus 42 (split by two) sites. In the non-IID case, however, the performance worsens after the split. We also find that data-sparse sites contribute noisy gradient updates in the 42-site scenario. For example, learning the local weights from a small cohort (e.g., Site 14, N=51) would be much more difficult than from a larger cohort (e.g., Site 6, N =1510) after splitting multiple ways.

**Model disparities across age and sex**
We examine potential sources of data heterogeneity including site locations, age, and sex, and their potential impact on FL training and performance. We identify correct COVID predictions from a 3D model that is trained on the first 10 sites, then evaluated on the remaining sites. We then use a t-SNE embedding cluster (**Figure 5**) to visualize the distribution of features, stratified (color-coded) by site locations and age. While the projected features across sites seem homogeneous, clusters emerge from 15.1, 6.2, 8.2 datasets. When the features are binned across age, clustering is more prominent. These clusters indicate that within the COVID population, the feature footprints are dissimilar across site locations and age subgroups.

We also illustrate the standard deviation of correctly identified features across age and sex (**Figure 5 b**). We find that males tend to exhibit wider variations within the feature space compared to females. Perhaps the AI is more brittle at evaluating males with COVID or there is inherent heterogeneity in the male population relative to the female. In the older COVID group (≥ 55 years) however, this feature gap, between males and females, diminishes.

We then conduct subgroup FL training and show the results in **Figure 5 (c, d)**. First, we train only the female population across the federation sites, then the younger (<55 years) and the older (≥ 55 years) age groups. We assess the subgroup accuracies for a hold-out site over the federated rounds. We observe that convergence is fairly robust for all cases. However, we find that the test set accuracies are higher for the subgroups that identify with the training population. For example, training on females yields much better performance on the female-only hold-outs.

**DISCUSSION**

We present a proof-of-concept of FL on a large-scale, heterogenous medical dataset, targeting 3D Chest CT lung evaluation across global geographic centers. We compare baseline siloed and CDS training versus various FL strategies. We also identify challenges inherent in real world data, such as data imbalance, missing classes, disease variants and severity, and data acquisition techniques, which collectively contribute to the non-IID nature of our cohort. Even among the correctly identified patients, we find that the AI model interprets differences within the subgroups, such as age, sex, and geographic locations.

With siloed training, even with a large cohort-site (e.g., Site 6.2), model generalization at other sites remains poor, possibly due to variable disease phenotypes or severity across sites, data acquisition parameters, and/or intra-site class imbalances. We also find that larger cohort-sites do not always yield the highest performance. Pretraining can improve per site average model accuracies, although not reliably. Dreaming synthetic images from a GAN approach may off-set class imbalance relevant to both siloed and FL training. Synthetic data augmentation from a GAN trained on public dataset can also improve FL training by mitigating class imbalance across sites. It is also noteworthy that synthetic data augmentation can also improve siloed training.

Even on heterogeneously distributed medical data, FL strategies perform close to CDS training, as shown by per-site and average accuracies. All FL methods and CDS show high success, except for a tiny cohort-site (Site 14), where the COVID and PNA groups are similarly severe, opacifying nearly the entire lung. Also, the number of local epochs per round is an important FL hyperparameter, likely since higher number allow for longer exposure to individual site data before a given aggregation round. However, too many local epochs could lead to model and client drift [26] and potential overfitting on small cohort-sites. An extreme case of drift is an FL model with only 1 global reduction round and 100 epochs of local training, where a global model that is simply the average over all model weights of the individual sites could lead to disastrous accuracies across all sites. In our experiment, we find the most robust convergence using 1 to 4 epochs of local training per reduction round.

The FL-specific hyperparameter C, the fraction of the FL network active in a given round, regulates the variation of sites used during a given global weight update for one reduction round. In C=1, all sites contribute, but the larger cohort-sites would effectively cast a larger vote during the final weight update than the underrepresented sites. In the non-IID case, we posit this could be a limitation, as the final model could learn only those features relevant to the larger cohorts. However, if C were too low, the FL network could be severely underutilized.

Despite potential clinical applicability of CIIL and IIL, these FL alternatives remain dependent on the site-to-site path chosen for training (**Figure 3(b)**). Thus, they are likely more suitable for learning in ecosystems with more balanced, IID datasets, and less desirable in medicine where not only are the pre-existing, static data heterogeneous, new data continue to evolve. Another striking observation is the vulnerability of CIIL and IIL to per site data sizes. When a shared model passes over a small cohort-site (e.g., 100 training samples) after traversing a large cohort-site (e.g., >1000 samples), new, noisy learning from the small site would often corrupt features learned from the previous large site.

It is noteworthy several sources of data heterogeneity can induce bias in FL training. For example, we find that patients from US and Canada occupy similar feature space, and thus difficult to distinguish, than those from Middle East, possibly due to differences in disease phenotypes, genetics, environmental, or other exogenous factors (e.g., data acquisition techniques).  We also find quantitative differences among the age (< 55 vs. ≥ 55 years) and sex (male vs. female) subgroups within the feature space (**Figure 5**). While FL remains feasible regardless of such subgroup differences, our preliminary results suggest that training within such a subgroup, e.g., females only, can yield AI models that are more robust to that population.

There are several limitations to consider.  Although we show a robust FL performance on par with CDS, it would be helpful to better interpret why some sites perform weaker than others. Also, while FL allows hospitals to train models without sharing data, recent studies have shown that privacy can still be leaked [30]. Finally, the present study represents FL in a simulation environment.  Future studies that examine a real-world FL learning platform and deployment could clarify additional barriers to clinical translation, including model vulnerabilities and privacy leakage in the context of FL.

**METHODS**

**Dataset**. Our cohort includes 16 global sites [Sites 1-16, totaling 21 hospitals (Site 11 containing 6 hospital affiliates)] (**Table 1**) from North and South America, Asia, Europe, and North Africa, including public datasets from China [3] and Russia [28], that represent 10,273 unique patients (median age 55 years) with 1,555,661 million images.  The decimal annotations (e.g., 5.1, 5.2) represent two independent datasets from that site (e.g., Site 5) separated by different timepoints, and thus for the study purpose, sources of two separate datasets. The cohort from each site consist of one or more of the following: 1) RT-PCR confirmed Covid (+) lung disease (COVID); 2) Covid (-) community acquired pneumonia (PNA); 3) and normal lung. The COVID cohort recruitment spans January 2020 and May 2021, likely including the alpha-, delta-, and possibly other (sub)variants.  The PNA group comprises bacterial or viral PNA prior to 2020 Covid pandemic, and if during the pandemic, considered to have Covid (-) infectious origins based on symptoms, imaging, negative RT-PCR, and serology or laboratory cultures. Institutional review board (IRB) at participating hospitals approved this retrospective study with waiver of consent since 1) the research involves no more than minimal risk to the participants with confidentiality maintained; 2) the waiver will not adversely affect the rights and welfare of the participants; and 3) information learned during the study would not affect the treatment of participants. Patient demographics are summarized in **Table 1.**

All raw images except those from Sites 1 and 12 are in Digital Imaging and Communications in Medicine (DICOM) format. The majority of the DICOMs contain a dynamic range of pixel intensities consistent with a lung window. Each patient DICOM images are scaled to 256 x 256 pixels. A volumetric stack containing 24 slices are then extracted across the lung axial plane for each CT exam. During training, 24-slices are sampled by a randomly generated axial offset value. We do not standardize the range of Hounsfield Units across sites so that the model would learn from heterogeneous standardizations.

**Model Architectures.** We use Deep COVID DeteCT (DCD) based on I3D Inception [1] and a 2.5D MobileNet, as the baseline AI architectures for the siloed, CDS, and FL experiments (**Figure 1**). For synthesizing new examples, we use a U-Net ResNet for the GAN Generator. All model architectures are displayed in our website[1]. After testing the 2D and 3D ResNet models, we find that the 2D models yield similar performance to the 2D MobileNet. Therefore, we use MobileNet for our 2D experiments as it offers ultra-low memory and compute footprint, rapid model share during FL, and its future potential on a mobile platform.

**Training**. For pre-training and GAN-based augmentation, we use a large public dataset (1.1). The models evaluate the efficacy of FL for multi-class classification (COVID, PNA, vs. normal lung). Each site-specific dataset is split into 50% training and 50% hold-out sets for validation. We also reserve a hold-out test dataset (a data subset from Site 11) that never participate in train-validation (**Figure 3(c)).** We create 5 random splits with different seeds and replicate FL experiments to ensure model stability and that the findings here are representative. Although all performance numbers represent one split configuration (a random seed), we find only slight deviation in the average and weighted average accuracies. For siloed and CDS training strategies, each model is trained for a simple, fixed number of local epochs and reduction rounds with Adam Optimization [31] and cross-entropy loss. We find little evidence that training for longer than 50 epochs decreases the validation performance. Further, slight gains in accuracy can be made with longer training. However, due to data complexities and heterogeneity, we choose a fixed set of epochs regardless of the experiment type. We recognize marginal improvements can be made with early stopping using internal evaluation on validation splits inside the training set.

Training FL models entails N federated rounds. We first ensure that each site has an exact copy of the initial weights before training can proceed. On each round, a randomly selected subset of the site participants is selected to be active. This active set performs a fixed number of local epochs of training and shares its weights to a parameter server. This server aggregates all the weights from each site, performs a weighted average over all participating models, then broadcasts this new set of weights to all sites to be used for the next round. During the aggregation, we choose not to broadcast and aggregate optimizer-specific parameters. To train CIIL and IIL models, we first configure a site-to-site training path that indicates the order of sites used for training. Traversing any given site consists of 1 local epoch of that site. Transitioning to the next site means holding the state (final weights and optimizer) from the end of training of the previous site to the start of the next site.

---

[1] Code available upon request and released soon on github.

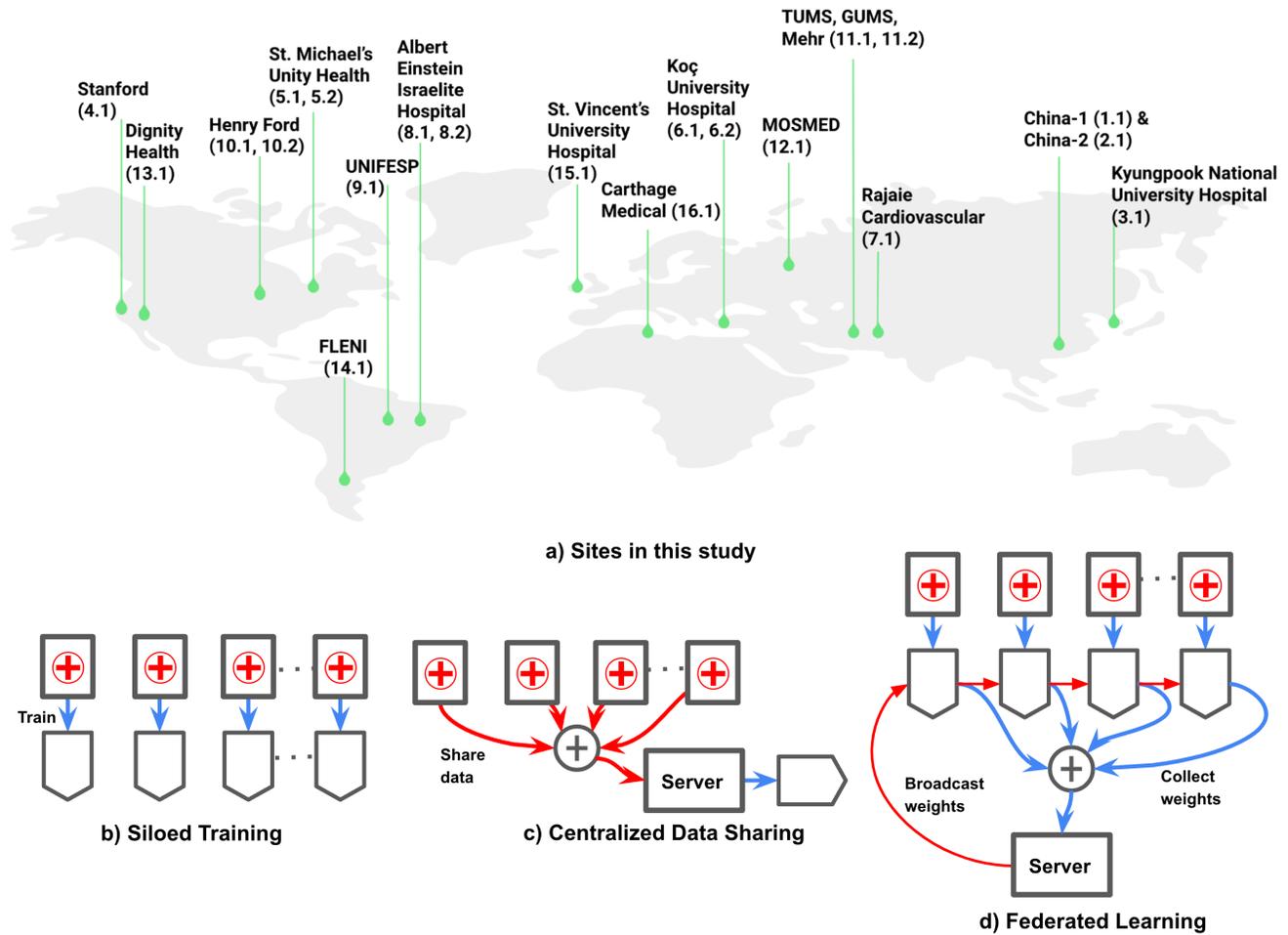

Figure 1. All sites used in this study (a). Three training methods investigated in this study (b,c,d).

Table 1. Demographic of sites. COVID- pneumonias (PNAs) and normal controls are also obtained. A few sites in Iran are merged due to the close proximity of hospitals. Several sites contained different recruitment times. For example, sites 11.1 and 11.2 differ in that data are collected from the same institution (TUMS) at two different times. TUMS is composed of many hospitals.

Table 1. Characteristics of Patients Included in Study by Institution and Country

| Institution | United States | | | Canada | Ireland | Brazil | | Argentina | Tunisia |
|---|---|---|---|---|---|---|---|---|---|
| | Stanford | Henry Ford | St. Joseph | Unity Health | St. Vincent | UNIFESP | Einstein | FLENI | Carthage |
| **COVID-19 Positive** | | | | | | | | | |
| Number of patients | 206 | 801 | 129 | 98 | 54 | 210 | 219 | 20 | 116 |
| Number of CT scans | 213 | 867 | 129 | 108 | 54 | 237 | 229 | 20 | 116 |
| Number of image slices | 39,854 | 192,639 | 14,437 | 20,720 | 14,613 | 33,869 | 88,058 | 2,486 | 31,155 |
| With contrast, no. (%) | 159 (75) | 627 (72) | 64 (50) | 28 (26%) | 52 (96%) | 4 (2) | 0 (0) | 0 (0) | 0 (0) |
| Age at initial CT scan, mean (SD), y | 52.5 (18.7) | 62.8 (16.4) | 54.8 (15.5) | 59 (14.8) | 61.6 (14.2) | 58.3 (15.4)[a] | 54.2 (14.1) | 61.4 (21.8) | 60.2 (14.1) |
| Female sex, no. (%) | 90 (44) | 408 (51) | 52 (40) | 39 (40) | 29 (54%) | 31 (35)[a] | 95 (43) | 11 (55) | 47 (41) |
| Deceased, no. (%) | 14 (7) | 73 (18)[a] | n/a | 1 (6)[a] | n/a | 21 (27)[a] | 1 (0) | n/a | n/a |
| **COVID-19 Negative** | | | | | | | | | |
| Number of patients | 41 | 218 | 101 | 77 | 15 | 41 | 38 | 20 | 0 |
| Number of CT scans | 41 | 218 | 101 | 77 | 15 | 41 | 38 | 20 | 0 |
| Number of image slices | 8,507 | 53,662 | 12,041 | 17,873 | 4,789 | 4,639 | 12,942 | 4,220 | 0 |
| With contrast, no. (%) | 19 (46) | 113 (52) | 8 (8) | 20 (26) | 10 (67%) | 2 (5) | 1 (3) | 0 (0) | - |
| Age at initial CT scan, mean (SD), y | 52.2 (18.1) | 62.2 (16.9) | 57.9 (16.1) | 60.8 (24.4) | 69.6 (13.8) | 52.6 (15.5) | 47.7 (31.8) | 47.7 (23.9) | - |
| Female sex, no. (%) | 19 (46) | 125 (57) | 38 (38) | 32 (42) | 6 (40%) | 19 (46) | 21 (55) | 10 (50) | - |
| **Normal Controls** | | | | | | | | | |
| Number of patients | 148 | 72 | 42 | 60 | 22 | 42 | 60 | 11 | 87 |
| Number of CT scans | 148 | 72 | 42 | 60 | 22 | 42 | 60 | 11 | 87 |
| Number of image slices | 30,708 | 17,803 | 5,045 | 13,739 | 7,517 | 4,524 | 18,098 | 1,953 | 22,191 |
| With contrast, no. (%) | 90 (61) | 48 (67) | 30 (71) | 34 (57) | 16 (73%) | 0 (0) | 0 (0) | 0 (0) | 0 (0) |
| Age at initial CT scan, mean (SD), y | 51.0 (16.2) | 42.8 (16.7) | 36.2 (13.1) | 50.5 (13.1) | 51.8 (18.1) | 56.4 (17.2) | 45.1 (15.9) | 41.7 (17.4) | 38.5 (14.4) |
| Female sex, no. (%) | 82 (55) | 41 (57) | 22 (52) | 31 (52) | 15 (68%) | 25 (60) | 27 (45) | 8 (73) | 53 (61) |

Table 1 (continued)

| Institution | Iran | | | | Turkey | S. Korea | China | | Russia |
|---|---|---|---|---|---|---|---|---|---|
| | TUMS | GUMS | Mehr | Rajaie | Koç | Kyungpook | China-1 | China-2 | MosMedData[c] |
| **COVID-19 Positive** | | | | | | | | | |
| Number of patients | 977 | 323 | 33 | 24 | 1219 | 60 | 964 | 103 | 856 |
| Number of CT scans | 997 | 323 | 33 | 24 | 1328 | 60 | 1,544 | 215 | 856 |
| Number of image slices | 261,020 | 9,579 | 3,059 | 4,622 | 254,532 | 19,701 | 37,056[b] | 39,630 | 20,544[b] |
| With contrast, no. (%) | 6 (1) | 0 (0) | 0 (0) | 0 (0) | 40 (3) | 11 (60) | 0 (0) | 1 (0) | 0 (0) |
| Age at initial CT scan, mean (SD), y | 57.3 (16.8) | 55.7 (17.1)[a] | 51.8 (17.3) | 54.9 (17.2) | 57.2 (15.9) | 65.1 (14.1) | n/a | 43.0 (15.1) | n/a |
| Female sex, no. (%) | 374 (38) | 159 (49)[a] | 10 (30) | 13 (54) | 532 (44) | 26 (43) | n/a | 40 (39) | n/a |
| Deceased, no. (%) | 172 (31)[a] | n/a | 1 (3) | 4 (17) | 24 (2) | 10 (17) | n/a | n/a | n/a |
| **COVID-19 Negative** | | | | | | | | | |
| Number of patients | 0 | 0 | 0 | 14 | 143 | 0 | 929 | 103 | 0 |
| Number of CT scans | 0 | 0 | 0 | 14 | 143 | 0 | 1,545 | 103 | 0 |
| Number of image slices | 0 | 0 | 0 | 1,801 | 36103 | 0 | 37,080[b] | 39,413 | 0 |
| With contrast, no. (%) | - | - | - | 0 (0) | 0 (0) | - | 0 (0) | 0 (0) | - |
| Age at initial CT scan, mean (SD), y | - | - | - | 50.3 (20.4) | 41.9 (14.1) | - | n/a | 39.0 (15.8) | - |
| Female sex, no. (%) | - | - | - | 5 (36) | 74 (52) | - | n/a | 45 (44) | - |
| **Normal Controls** | | | | | | | | | |
| Number of patients | 145 | 61 | 0 | 23 | 148 | 0 | 850 | 96 | 254 |
| Number of CT scans | 145 | 61 | 0 | 23 | 148 | 0 | 1,078 | 100 | 254 |
| Number of image slices | 32,971 | 1,821 | 0 | 3,253 | 31365 | 0 | 25,872[b] | 12,061 | 6,096[b] |
| With contrast, no. (%) | 0 (0) | 0 (0) | - | 0 (0) | 0 (0) | - | 0 (0) | 0 (0) | 0 (0) |
| Age at initial CT scan, mean (SD), y | 35.5 (10.4) | 36.2 (11.9) | - | 34.3 (10.9) | 32.1 (7.5) | - | n/a | 35.2 (5.8) | n/a |
| Female sex, no. (%) | 68 (47) | 35 (57) | - | 7 (30) | 90 (61) | - | n/a | 58 (60) | n/a |

Abbreviation: CT, computerized tomography; GUMS, Guilan University of Medical Sciences; SD, standard deviation; TUMS, Tehran University of Medical Sciences; UNIFESP, Universidade Federal de São Paulo; [a] Based on limited subset due to gaps in data; [b] After pre-processing scans to 24 slices each; [c] Complete dataset includes 622 females (56%) and 488 males (44%) with a median age of 47, ranging from 18 to 97 years

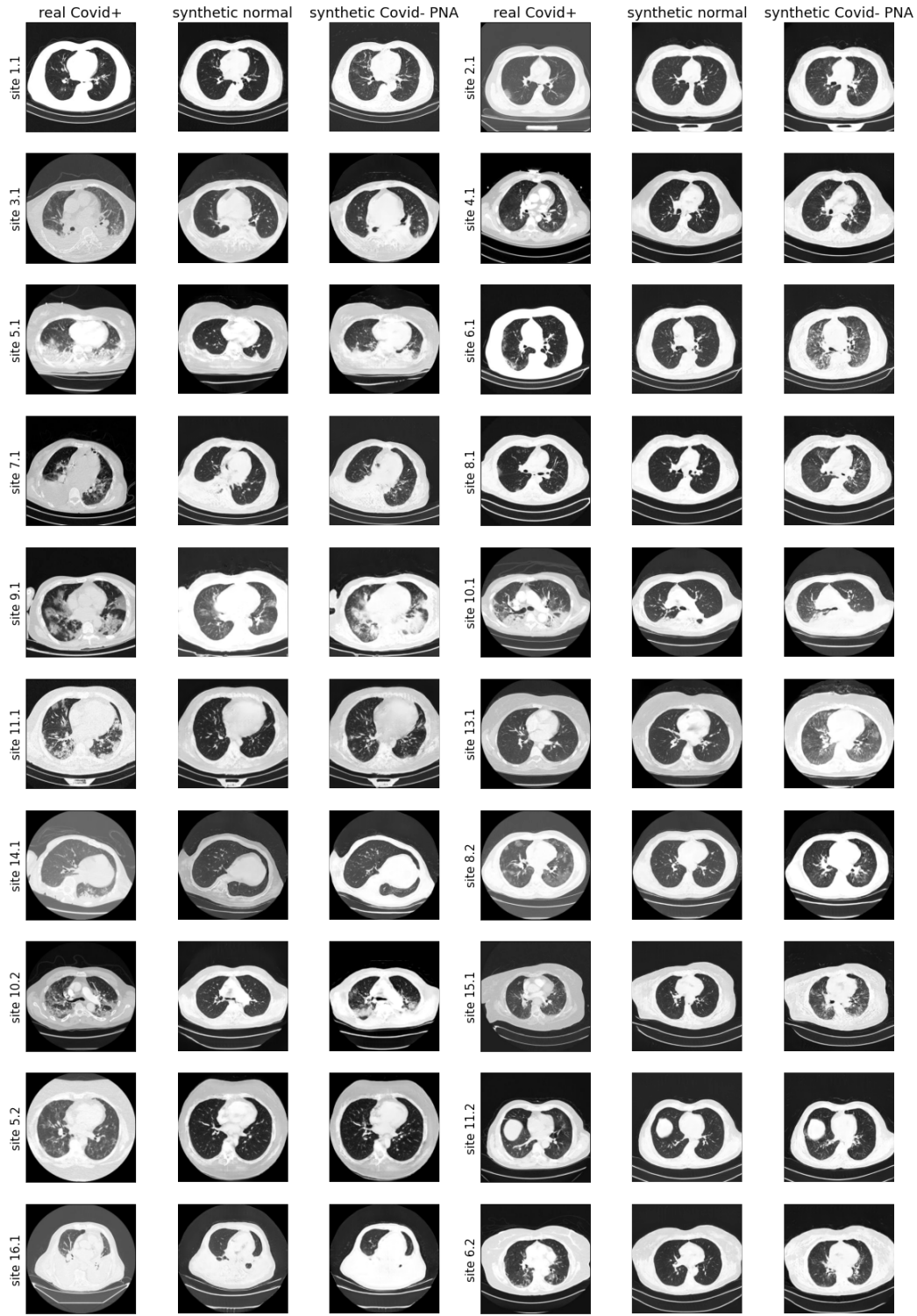

Figure 2. Illustration of the synthetic data generated per site and diversity of the dataset. Examples of the site-specific synthetic images (2D slice of a 3D image) of other classes generated on real patient scans (with COVID) from 20 selected sites. Synthetic data is used to improve non-FL and FL training. These synthetic images are only used in the training process and not for validation. The disease severity was found to be a strong function of the site location.

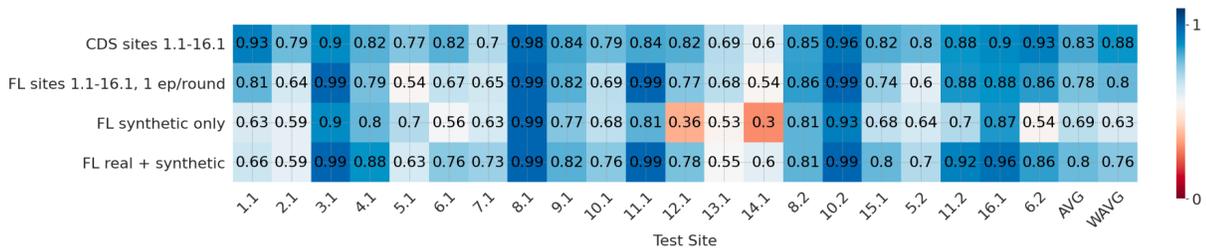

(a)

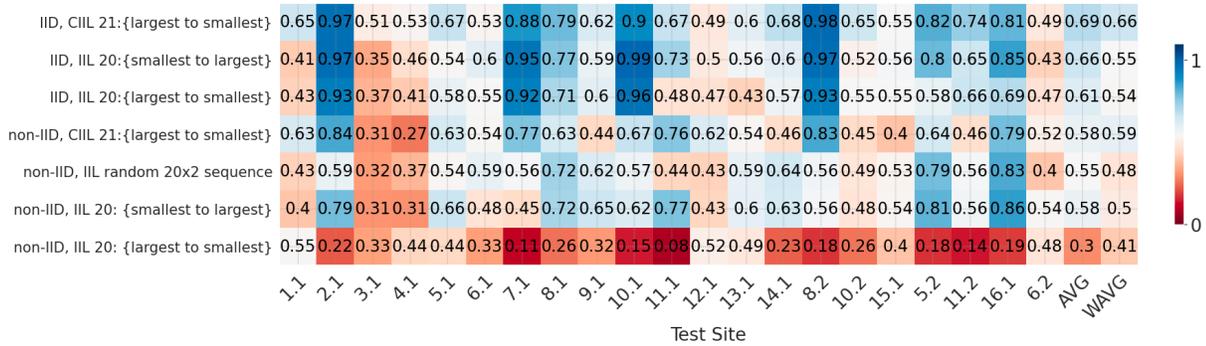

(b)

| Methods | Accuracy on hold-out validation sets | | | Accuracy on additional hold-out set |
|---|---|---|---|---|
| | WAVG | AVG | F1-score | |
| Best siloed | 0.65 | 0.58 | 0.62 | 0.69 |
| Centralized Data Sharing (CDS) | 0.83 | 0.88 | 0.84 | 0.93 |
| FL (1 local epoch per round) | 0.78 | 0.80 | 0.79 | 0.86 |
| FL (4 local epoch per round) | 0.80 | 0.84 | 0.80 | 0.85 |
| FL on synthetic data only | 0.69 | 0.63 | 0.70 | 0.65 |
| FL on real + synthetic | 0.80 | 0.76 | 0.79 | 0.83 |

(c)

Figure 3. Accuracies using FL strategies with pretraining and synthetic data generation (a) and other collaborative training methods (b). In (b), different training path configurations using IIL And CIIL methods are shown; each site visit consists of 1 epoch on the training set of that site. The bottom row represents the site accuracies after training with IIL from the largest site (excluding site 1.1) to the smallest site. The reverse path (smallest to largest) is shown in the row just above. The inclusion of site 1.1 leads to very modest performance gains, especially for the largest to smallest path. The fourth row from the bottom represents CIIL training with 3 rounds starting with the largest site (1) to the smallest. The top row represents CIIL training with artificially-generated IID partitioning on the training sites with 3 rounds. Table (c) showing summary of FL and non-FL performance.

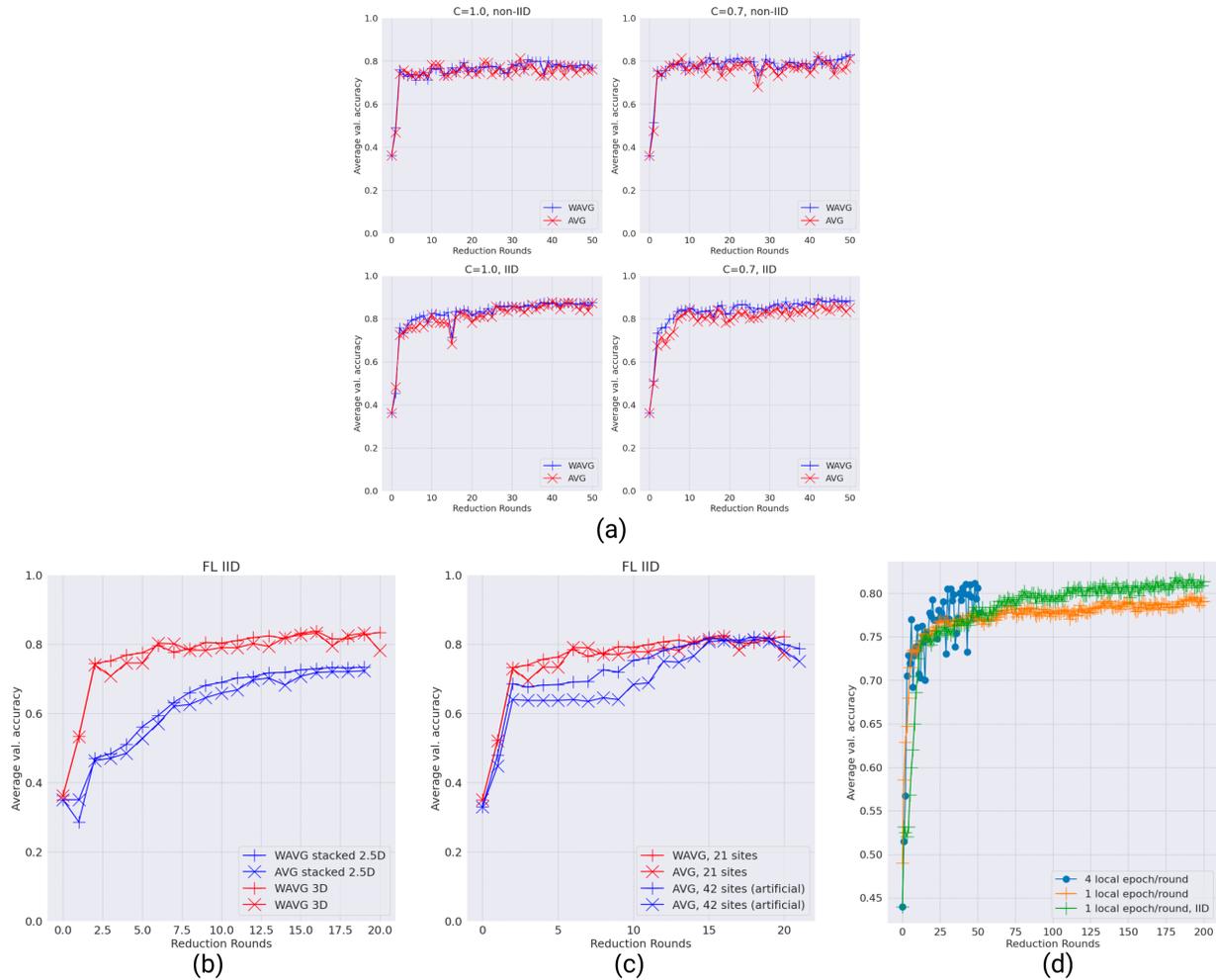

Figure 4. Different configurations of FL parameters. FL training for non-IID and IID splits (a) for two different *C* (fraction of active clients per round). The non-IID case is the default scenario with sites representing the data collected at one hospital for a given collection time. The IID case is generated by aggregating data from all sites, splitting into 21 equal partitions, and assigning each to a unique site ID. Comparison of 2D and 3D models (b). Convergence where each of the original 21 sites are artificially split up into 2 disjoint sites (c). Training convergence of runs for a total of 200 epochs: one with 4 local epochs per round and two with 1 local epoch per round for non-IID (realistic) and IID sites (d).

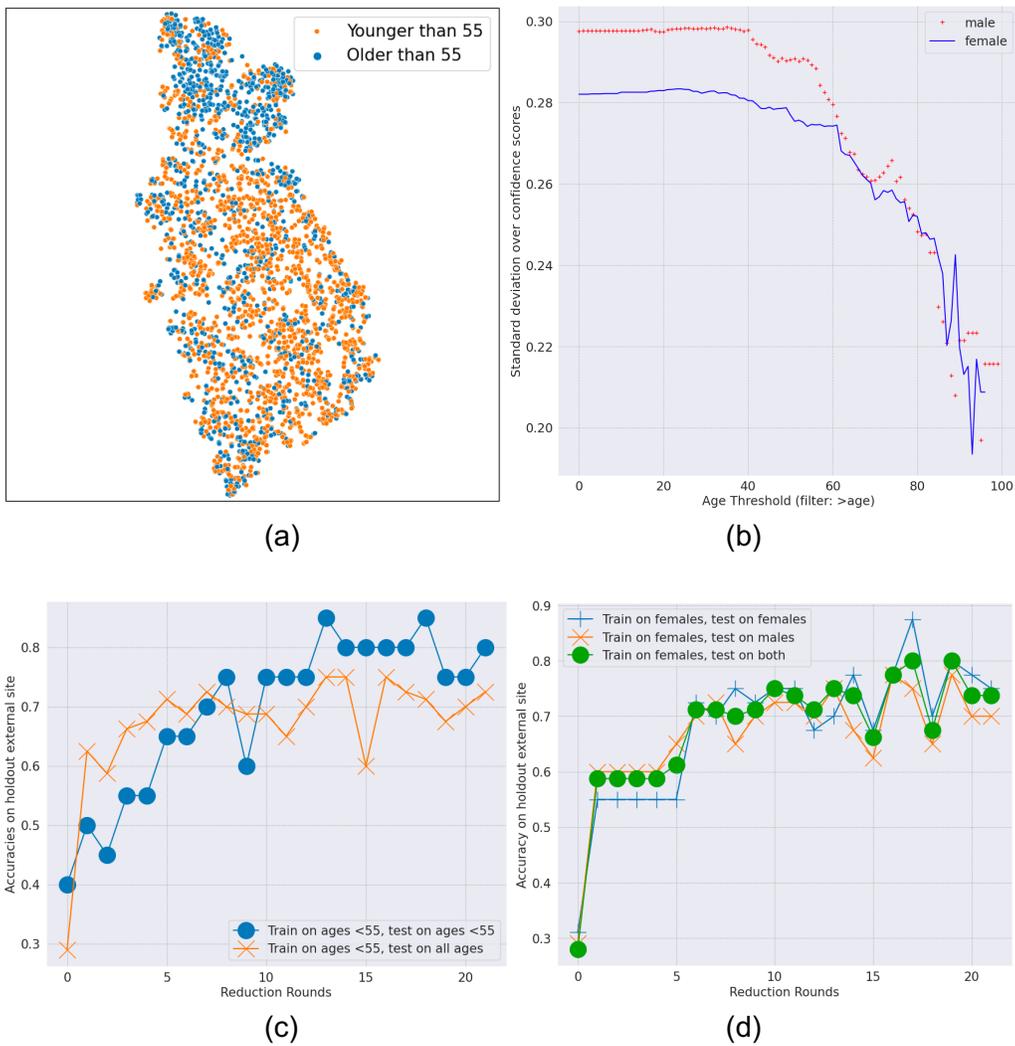

Figure 5. t-SNE embeddings with color-coded age and demographic clusters (a). The variations in feature space for males and females and across different age groups (b). FL evaluation on a hold-out site when performing FL training across sex and age subgroups (c, d).

Supplemental Figures

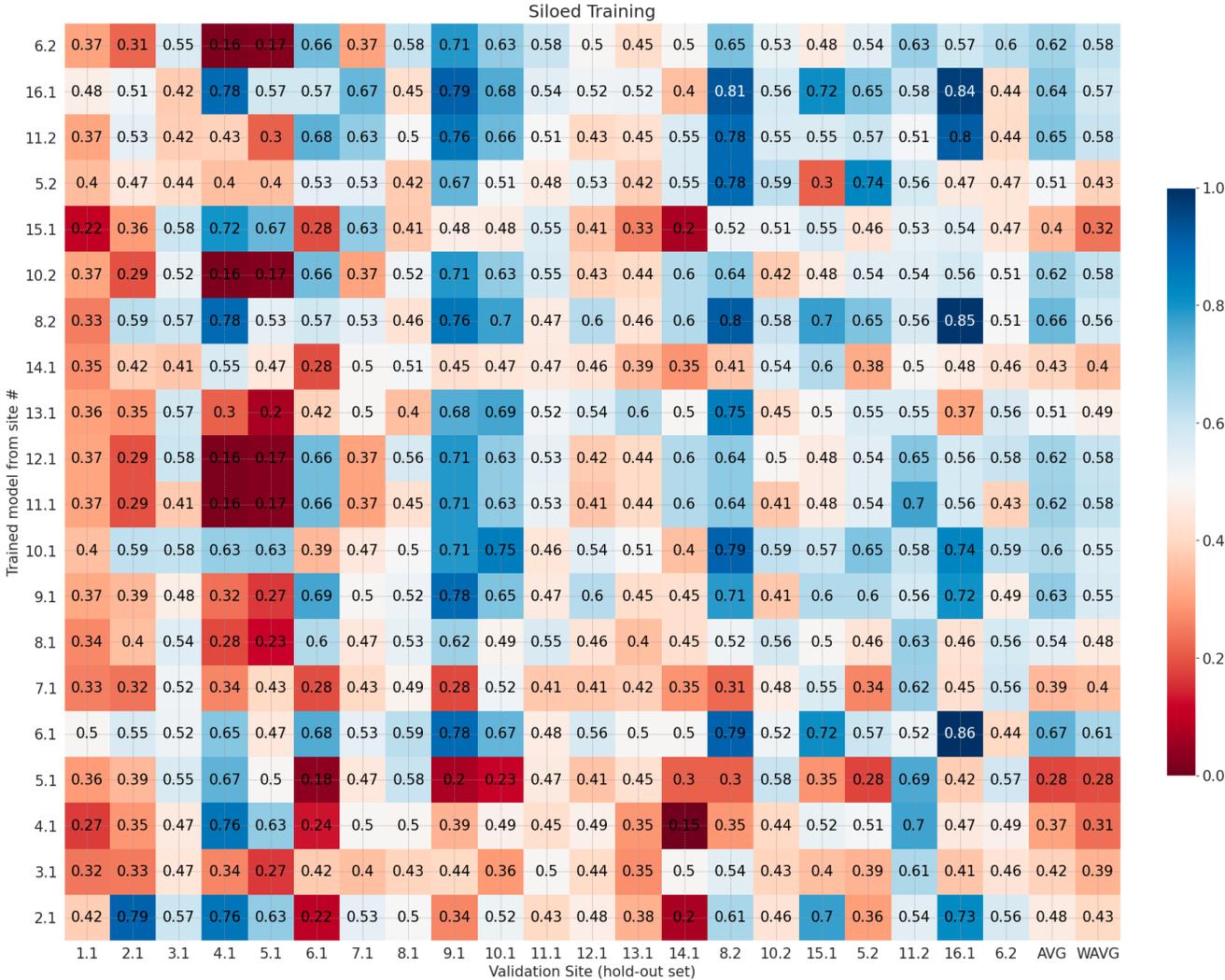

Figure S1(a). Quality of single site strategies. The matrices represent the (hold-out) validation accuracies of 21 sites using 20 independently-trained models on siloed data. The average accuracies (AVG) and weighted average accuracies (WAVG), weighted by the data size per site, are shown.

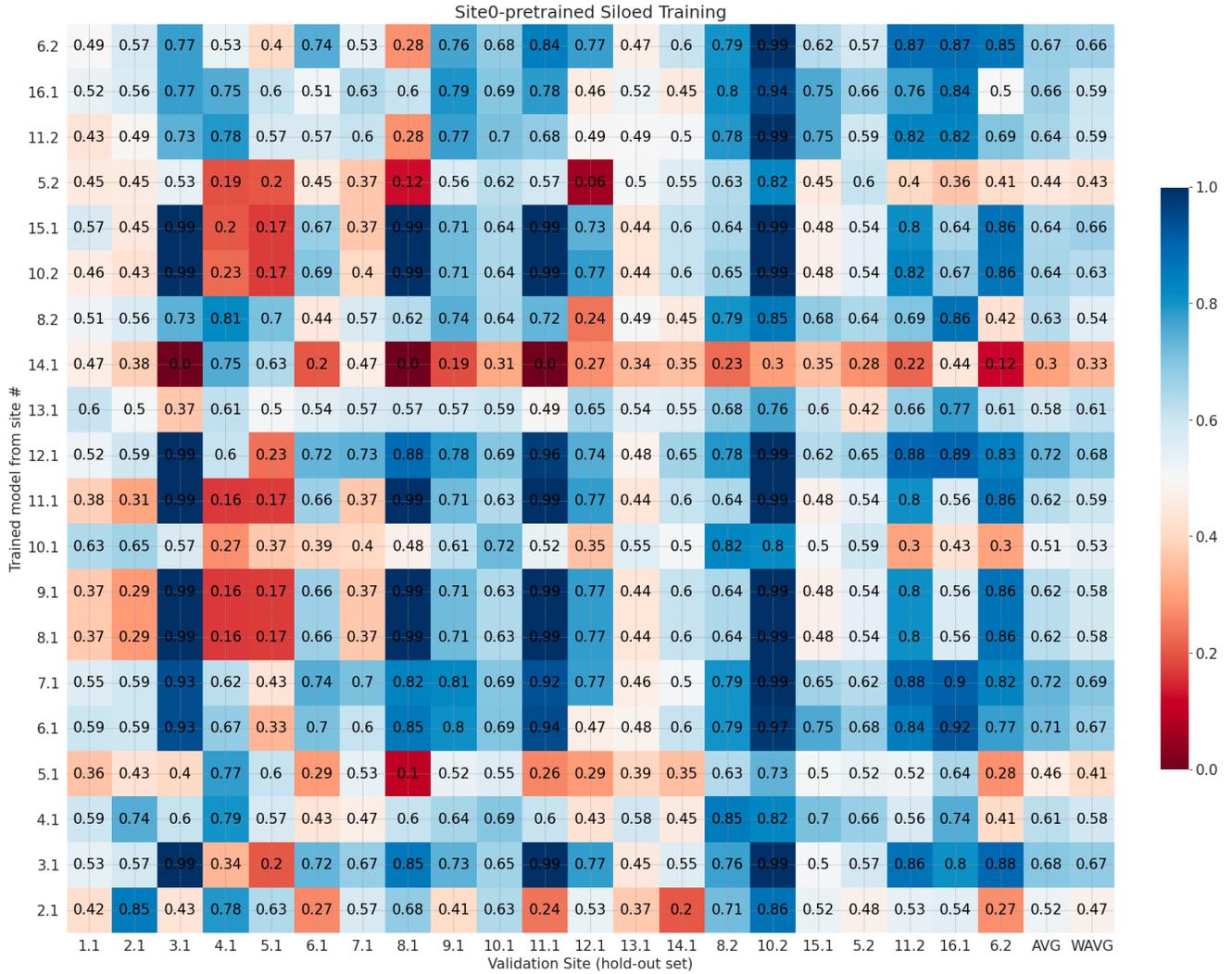

Figure S1(b). Quality of single site strategies with pretraining on site 0. The matrices represent the (hold-out) validation accuracies of 21 sites using 20 independently-trained models on siloed data. The average accuracies (AVG) and weighted average accuracies (WAVG), weighted by the data size per site, are shown.

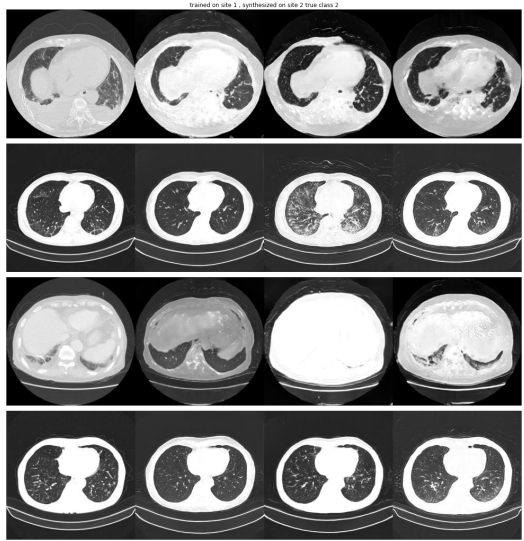

Figure S2. Examples of difficult synthetic images from 4 different sites (left). The first column represents real images, second column synthesized normals, third column synthesized PNAs, and fourth synthesized COVID+. Real examples of COVID- PNA patients in our study that show heterogeneous features, some that are similar to COVID+ PNA (right). For example, two COVID- patients with influenza PNA (arrows), including H1N1 (long arrow), show peripheral ground glass opacities similar to COVID+.